\documentclass[runningheads]{llncs}

\usepackage{eccv}

\usepackage{eccvabbrv}
\usepackage{graphicx}
\usepackage{booktabs}
\usepackage[accsupp]{axessibility}
\usepackage{hyperref}
\usepackage{orcidlink}

\usepackage{amsmath,amsfonts,bm}

\def\eqref#1{equation~\ref{#1}}

\def\1{\bm{1}}

\def\vc{{\bm{c}}}

\def\vp{{\bm{p}}}

\def\vs{{\bm{s}}}

\def\vv{{\bm{v}}}

\def\vz{{\bm{z}}}

\def\mC{{\bm{C}}}

\def\mP{{\bm{P}}}

\DeclareMathAlphabet{\mathsfit}{\encodingdefault}{\sfdefault}{m}{sl}
\SetMathAlphabet{\mathsfit}{bold}{\encodingdefault}{\sfdefault}{bx}{n}

\usepackage{microtype}
\usepackage{todonotes}
\usepackage{url}
\usepackage{multirow}
\usepackage{pifont}
\usepackage{xcolor, colortbl}
\usepackage{xspace}
\usepackage{amsmath}
\usepackage{amssymb}
\usepackage{mathtools}
\usepackage{pgfplots}
\usepackage{wrapfig}
\usepackage{adjustbox}

\pgfplotsset{compat=1.18}

\usepackage{algorithm}
\usepackage{algorithmic}

\usepackage[capitalize,noabbrev]{cleveref}
\crefname{figure}{Fig.}{Figs.}
\crefname{equation}{Eq.}{Eqs.}

\newcommand{\cmark}{\ding{51}}
\newcommand{\xmark}{\ding{55}}
\newcommand{\ours}{ExCB\xspace}

\begin{document}

\title{Efficient Unsupervised Visual Representation Learning with Explicit Cluster Balancing}
\titlerunning{Unsupervised Representation Learning with Explicit Cluster Balancing}

\author{Ioannis Maniadis Metaxas\inst{1}\thanks{Corresponding author.} \and
Georgios Tzimiropoulos\inst{1} \and
Ioannis Patras\inst{1}}

\authorrunning{I.~Maniadis Metaxas, G.~Tzimiropoulos, I.~Patras}

\institute{Queen Mary University of London \\ \email{\{i.maniadismetaxas,g.tzimiropoulos,i.patras\}@qmul.ac.uk}}

\maketitle

\begin{abstract}
Self-supervised learning has recently emerged as the preeminent pretraining paradigm across and between modalities, with remarkable results. 
In the image domain specifically, group (or cluster) discrimination has been one of the most successful methods.
However, such frameworks need to guard against heavily imbalanced cluster assignments to prevent collapse to trivial solutions.
Existing works typically solve this by reweighing cluster assignments to promote balance, or with offline operations (e.g. regular re-clustering) that prevent collapse.
However, the former typically requires large batch sizes, which leads to increased resource requirements, and the latter introduces scalability issues with regard to large datasets.
In this work, we propose \ours, a framework that tackles this problem with a novel cluster balancing method.
\ours estimates the relative size of the clusters across batches and balances them by adjusting cluster assignments, proportionately to their relative size and in an online manner.
Thereby, it overcomes previous methods' dependence on large batch sizes and is fully online, and therefore scalable to any dataset.
We conduct extensive experiments to evaluate our approach and demonstrate that \ours: \textbf{a)} achieves state-of-the-art results with significantly reduced resource requirements compared to previous works, \textbf{b)} is fully online, and therefore scalable to large datasets, and \textbf{c)} is stable and effective even with very small batch sizes.
Code and models will be made available \href{https://github.com/ManiadisG/ExCB}{here}.
\end{abstract}

\setlength{\tabcolsep}{10pt}

\section{Introduction}

\begin{figure}[t]
\begin{center}
\includegraphics[width=1.\linewidth]{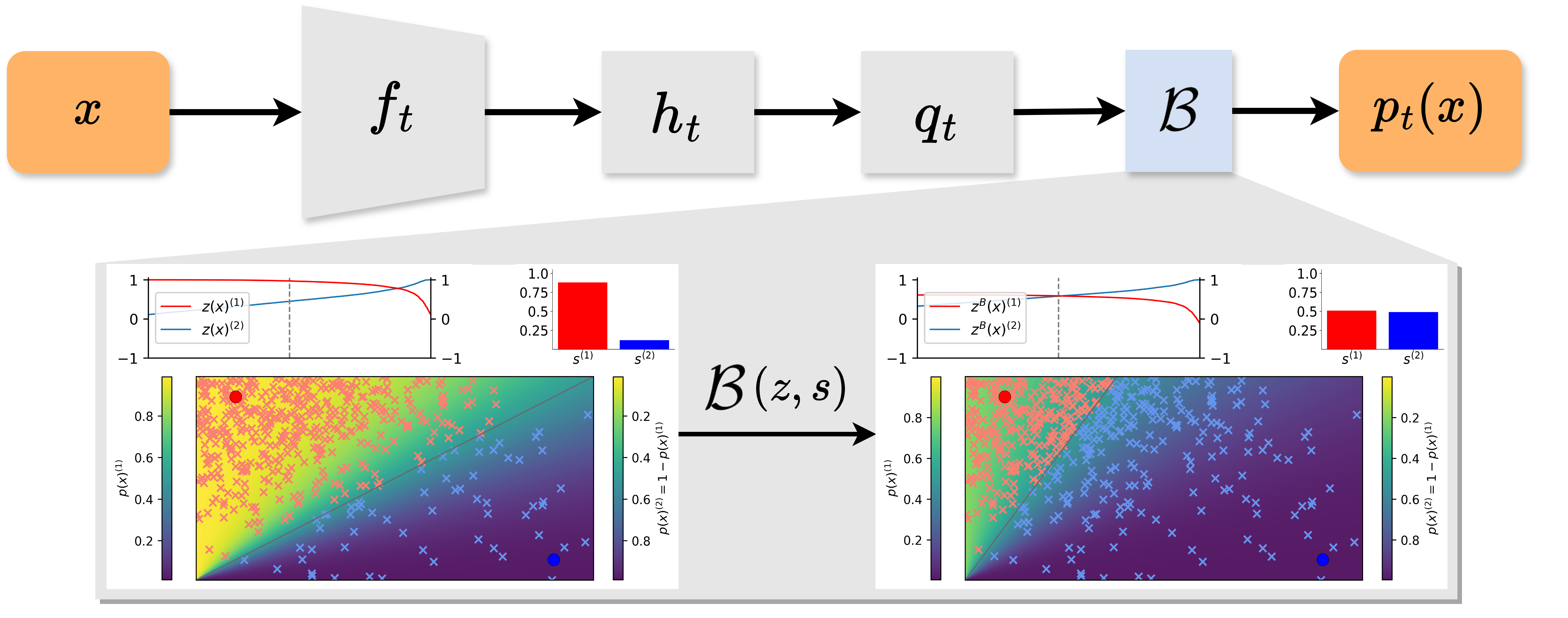}
\caption{Illustration of \ours's balancing operator $\mathcal{B}$ for two clusters $c_1$ (red) and $c_2$ (blue). $\mathcal{B}(z;s)$ adjusts sample-cluster cosine similarities $z$ according the relative cluster sizes, as measured in $s$. For smaller clusters the similarities are increased ($z^B>z$), whereas for larger clusters the similarities are decreased ($z^B<z$). The impact, as seen in the figure, is that the boundary between clusters shifts, undersized (oversized) clusters are assigned more (fewer) samples, and clusters become more balanced.}
\label{fig:main}
\end{center}
\end{figure}

Unsupervised, or self-supervised, representation learning has recently emerged as a dominant training paradigm to leverage vast amounts of unlabeled data in order to train powerful models.
That is typically done by utilizing natural supervisory signals inherent in each domain~\cite{devlin2018bert,pirl,kostas2021bendr,niizumi2021byol,baevski2020wav2vec} or between domains~\cite{radford2021learning,afouras2020self,baevski2022data2vec} in order to craft training objectives, thereby overcoming the need for extensive training data annotations.
In the visual domain, various types of self-supervised objectives have been proposed, such as transformation prediction~\cite{rotnet}, reconstruction~\cite{mae}, instance discrimination~\cite{simclr,mocov2} and group discrimination~\cite{deepcluster}.

Group (or cluster) discrimination methods train models with pseudo-labels (or cluster assignments), so that representations from similar images (that are assigned to the same cluster) are pulled together, and different images (assigned to different clusters) are pushed away.
This approach has been very effective, as it alleviates the false-negatives issue that plagues instance discrimination-based methods~\cite{triplet}.
A major distinction within clustering-based methods is whether the pseudo-labels are produced online or offline.
Offline methods~\cite{deepcluster,sela} periodically operate on the entire dataset to cluster the data and produce pseudo-labels and/or centroids. These methods are straightforward and stable, but cannot easily scale to large datasets, as they require storing and operating on features from the entire dataset.
Online methods~\cite{dino,mira,swav,coke,twist} produce pseudo-labels for each batch during training and do not require access to the entire dataset at any one time. They can, therefore, scale to any number of samples.
However, online methods need to apply balancing constraints to the cluster assignments to avoid collapse (i.e. all samples being assigned to the same cluster).
Typically, these constraints are applied using in-batch statistics: cluster assignments within each batch are adjusted to approximate balanced clusters, often by solving optimization problems~\cite{swav,mira}.
This approach requires large batches to work effectively, as the assumption of in-batch balance stands only if the number of samples in the batch is comparable to the number of clusters. In turn, this leads to increased resource requirements and, possibly, training instability.

In this work we propose \ours, a self-supervised framework that utilizes a novel online cluster balancing method and achieves state-of-the-art performance with remarkable efficiency in terms of training time and batch size requirements.
\ours relies on two simple components: 
\textbf{i)} We measure the relative size of clusters over multiple steps using hard assignments. Thereby, \ours approximates the cluster distribution across the dataset accurately and reliably, without depending on volatile in-batch statistics and the assignments' confidence, and without requiring a large batch size.
\textbf{ii)} We adjust cluster assignments according to each cluster's size, as measured in \textbf{(i)}. Specifically, we increase (decrease) the sample-cluster similarities of small (large) clusters to assign them more (fewer) samples, thereby ensuring that no cluster deviates significantly in terms of their size in either direction.
Together, the two components enforce an explicit soft balancing constraint on clusters that: 
\textbf{a)} is fully online without requiring a large batch size,
\textbf{b)} has negligible computational cost, as it does not require solving optimization problems,
and \textbf{c)} is, conceptually, very intuitive and straightforward.

\definecolor{gainsboro}{rgb}{0.86, 0.86, 0.86}
\definecolor{OceanBlue}{RGB}{0, 102, 204}
\definecolor{ForestGreen}{RGB}{34, 139, 34}
\definecolor{SunnyYellow}{RGB}{255, 204, 0}
\definecolor{RubyRed}{RGB}{215, 48, 39}
\definecolor{PurplePlum}{RGB}{123, 50, 148}
\definecolor{TurquoiseTeal}{RGB}{0, 128, 128}
\definecolor{Goldenrod}{RGB}{218, 165, 32}
\definecolor{SoftOrange}{RGB}{255, 165, 79}
\definecolor{CharcoalGray}{RGB}{64, 64, 64}
\definecolor{Crimson}{RGB}{220, 20, 60}

\newcommand{\coloredcircle}[1]{\textcolor{#1}{\rule{8pt}{8pt}}}

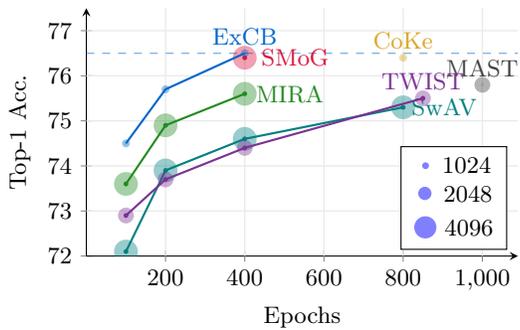
\begin{wrapfigure}{r}{0.55\textwidth}
    \centering
\hspace{-0.3cm}
\begin{tikzpicture}

        \begin{axis}[
            xlabel={Epochs},
            ylabel={Top-1 Acc.},
            xmin=0, xmax=1090,
            ymin=72.0, ymax=77.5,
            xtick={200, 400, 600, 800, 1000},
            ytick={72, 73, 74, 75, 76, 77},
            axis lines=left,
            legend style={
            at={(0.975,0.025)}, 
            anchor=south east, 
            font=\footnotesize,
            legend cell align=left,
            },
            width=0.6\textwidth,
            height=0.4\textwidth,
        ]

        \draw[gainsboro!60, line width=0.5pt] (axis cs:200, \pgfkeysvalueof{/pgfplots/ymin}) -- (axis cs:200, \pgfkeysvalueof{/pgfplots/ymax});
        \draw[gainsboro!60, line width=0.5pt] (axis cs:400, \pgfkeysvalueof{/pgfplots/ymin}) -- (axis cs:400, \pgfkeysvalueof{/pgfplots/ymax});
        \draw[gainsboro!60, line width=0.5pt] (axis cs:600, \pgfkeysvalueof{/pgfplots/ymin}) -- (axis cs:600, \pgfkeysvalueof{/pgfplots/ymax});
        \draw[gainsboro!60, line width=0.5pt] (axis cs:800, \pgfkeysvalueof{/pgfplots/ymin}) -- (axis cs:800, \pgfkeysvalueof{/pgfplots/ymax});
        \draw[gainsboro!60, line width=0.5pt] (axis cs:1000, \pgfkeysvalueof{/pgfplots/ymin}) -- (axis cs:1000, \pgfkeysvalueof{/pgfplots/ymax});
        \draw[gainsboro!60, line width=0.5pt] (axis cs:\pgfkeysvalueof{/pgfplots/xmin}, 73) -- (axis cs:\pgfkeysvalueof{/pgfplots/xmax}, 73);
        \draw[gainsboro!60, line width=0.5pt] (axis cs:\pgfkeysvalueof{/pgfplots/xmin}, 74) -- (axis cs:\pgfkeysvalueof{/pgfplots/xmax}, 74);
        \draw[gainsboro!60, line width=0.5pt] (axis cs:\pgfkeysvalueof{/pgfplots/xmin}, 75) -- (axis cs:\pgfkeysvalueof{/pgfplots/xmax}, 75);
        \draw[gainsboro!60, line width=0.5pt] (axis cs:\pgfkeysvalueof{/pgfplots/xmin}, 76) -- (axis cs:\pgfkeysvalueof{/pgfplots/xmax}, 76);
        \draw[gainsboro!60, line width=0.5pt] (axis cs:\pgfkeysvalueof{/pgfplots/xmin}, 77) -- (axis cs:\pgfkeysvalueof{/pgfplots/xmax}, 77);
        \draw[dashed, OceanBlue, line width=0.5pt, draw opacity=0.5] (axis cs:\pgfkeysvalueof{/pgfplots/xmin}, 76.5) -- (axis cs:\pgfkeysvalueof{/pgfplots/xmax}, 76.5);

        \footnotesize{

        \coordinate (G1) at (1000,75.8);
        \draw[CharcoalGray, thick] (G1) circle (0.01pt);
        \fill[CharcoalGray, opacity=0.4] (G1) circle (3pt);
        \node[above] at (G1) {\textcolor{CharcoalGray}{MAST}};
        
        \coordinate (F1) at (800,76.4);
        \draw[Goldenrod, thick] (F1) circle (0.01pt);
        \fill[Goldenrod, opacity=0.4] (F1) circle (1.5pt);
        \node[above] at (F1) {\textcolor{Goldenrod}{CoKe}};
        
        \coordinate (E1) at (100,72.1); 
        \coordinate (E2) at (200,73.9); 
        \coordinate (E3) at (400,74.6); 
        \coordinate (E4) at (800,75.3);
        \draw[TurquoiseTeal, thick] (E1) circle (0.5pt) -- (E2) circle (0.5pt) -- (E3) circle (0.5pt) -- (E4) circle (0.5pt);
        \fill[TurquoiseTeal, opacity=0.4] (E1) circle (4.5pt);
        \fill[TurquoiseTeal, opacity=0.4] (E2) circle (4.5pt);
        \fill[TurquoiseTeal, opacity=0.4] (E3) circle (4.5pt);
        \fill[TurquoiseTeal, opacity=0.4] (E4) circle (4.5pt);
        \node[right] at (E4) {\textcolor{TurquoiseTeal}{SwAV}};
        
        \coordinate (D1) at (100,72.9); 
        \coordinate (D2) at (200,73.7); 
        \coordinate (D3) at (400,74.4); 
        \coordinate (D5) at (850,75.5);
        \draw[PurplePlum, thick] (D1) circle (0.5pt) -- (D2) circle (0.5pt) -- (D3) circle (0.5pt) -- (D5) circle (0.5pt);
        \fill[PurplePlum, opacity=0.4] (D1) circle (3pt);
        \fill[PurplePlum, opacity=0.4] (D2) circle (3pt);
        \fill[PurplePlum, opacity=0.4] (D3) circle (3pt);
        \fill[PurplePlum, opacity=0.4] (D5) circle (3pt);
        \node[above] at (D5) {\textcolor{PurplePlum}{TWIST}};
        
        \coordinate (C1) at (100,73.6); \coordinate (C2) at (200,74.9); \coordinate (C3) at (400,75.6);
        \draw[ForestGreen, thick] (C1) circle (0.5pt) -- (C2) circle (0.5pt) -- (C3) circle (0.5pt);
        \fill[ForestGreen, opacity=0.4] (C1) circle (4.5pt);
        \fill[ForestGreen, opacity=0.4] (C2) circle (4.5pt);
        \fill[ForestGreen, opacity=0.4] (C3) circle (4.5pt);
        \node[right] at (C3) {\textcolor{ForestGreen}{\hspace{1pt}MIRA}};
        
        \coordinate (B1) at (400,76.4);
        \draw[Crimson, thick, fill=Crimson] (B1) circle (0.5pt);
        \fill[Crimson, opacity=0.4] (B1) circle (4.5pt);
        \node[right] at (B1) {\textcolor{Crimson}{\ SMoG}};
        
        \coordinate (A1) at (100,74.5); \coordinate (A2) at (200,75.7); \coordinate (A3) at (400,76.5);
        \draw[OceanBlue, thick] (A1) circle (0.01pt) -- (A2) circle (0.01pt) -- (A3) circle (0.01pt);
        \fill[OceanBlue, opacity=0.4] (A1) circle (1.5pt);
        \fill[OceanBlue, opacity=0.4] (A2) circle (1.5pt);
        \fill[OceanBlue, opacity=0.4] (A3) circle (1.5pt);
        \node[above] at (A3) {\textcolor{OceanBlue}{\ours}};

        \addlegendimage{empty legend}
        \addlegendentry{\hspace{3pt}\raisebox{0.5\height}{\tikz{\node[circle, fill=blue, fill opacity=0.5, inner sep=0.5pt] {};}}\hspace{2pt} 1024}
        \addlegendimage{empty legend}
        \addlegendentry{\hspace{1.5pt}\raisebox{0.15\height}{\tikz{\node[circle, fill=blue, fill opacity=0.5, inner sep=1.4pt] {};}}\hspace{1.5pt} 2048}
        \addlegendimage{empty legend}
        \addlegendentry{\raisebox{-0.15\height}{\tikz{\node[circle, fill=blue, fill opacity=0.5, inner sep=2.6pt] {};}} 4096}
        }
        \end{axis}

\end{tikzpicture}
\caption{Linear classification accuracy on ImageNet with ResNet50 for different self-supervised methods. Circles indicate pretraining batch size. \ours achieves state-of-the-art results with the most efficient combination of few epochs and small batch size.}
\label{fig:results}
\end{wrapfigure}

We conduct extensive experiments and demonstrate that \ours achieves state-of-the-art results with both convolutional and transformer backbones. Importantly, that is achieved with a much smaller batch size and/or fewer training epochs compared to previous works. Finally, we show that, out of the box, \ours is stable and achieves strong performance even with very small batch sizes. Jointly, our findings show that \ours achieves state-of-the-art performance in self-supervised representation learning while also being remarkably efficient, which has the additional benefit of significantly lowering the resource-wise barrier of entry for self-supervised pretraining.

\section{Related Works}\label{sec:related_works}

\subsection{Self-supervised learning for visual data}

The objective of self-supervised learning is to leverage \textit{pretext} tasks in order to learn robust representations from non-annotated data, such that they can be used to effectively solve other, \textit{downstream} tasks with minimal supervision. Pretext tasks in the visual domain typically leverage strong data augmentations to formulate training objectives, such as identifying transformations~\cite{rotnet}, patch permutation~\cite{jigsawssl} and instance discrimination~\cite{pirl}. Recent self-supervised representation learning methods can be roughly grouped based on the objectives they use: a) instance-discrimination~\cite{mocov2,dcl,chen2021empirical,nnclr,relic,simclr,wmse,cgh,maskcon}, where samples' views must be discernible from other samples, b) non-contrastive methods, where transformation invariance is imposed via self-distillation~\cite{byol,simsiam,huang2023mast} or by enforcing specific properties to batch feature statistics~\cite{tico,vicreg,bardes2022vicregl,barlowtwins,sslhsic}, c) reconstruction-based methods~\cite{msn,mae,simmim}, where the model predicts missing image patches, and d) clustering-based methods, where supervision is provided by pseudo-labels, typically produced via clustering~\cite{deepcluster,swav,mira,dino,twist,coke,pcl,smog,obow}. 
We emphasize that this taxonomy is indicative, and is neither exact nor strict, as the various methods have individual features that could place them in more than one or in distinct groups. Furthermore, we note that, whereas self-supervised representation learning is primarily focused on classification-based downstream tasks, a distinct line of research is focused on self-supervised learning frameworks that prioritize dense prediction tasks such as detection and segmentation~\cite{wang2021dense,wen2022self,xie2021propagate,bardes2022vicregl,stegmuller2023croc,henaff2022object,henaff2021efficient}.

\subsection{Clustering-based self-supervised learning}

Self-supervised frameworks based on group discrimination generate sample pseudo-labels (typically via clustering) and use them as supervision to train the model.
A key distinction among these methods is whether they are online or offline.

Offline methods~\cite{deepcluster,pcl} typically produce pseudo-labels by regularly clustering samples across the dataset. Distinctly,~\cite{sela} produces pseudo-labels by regularly extracting cluster assignments for the entire dataset and, after balancing them with the Sinkhorn-Knopp algorithm~\cite{cuturi2013sinkhorn}, uses them to further train the model. This approach guarantees training stability but introduces additional computational cost. More importantly, however, offline methods are very hard to scale to large datasets, as they require the storing and processing (i.e. clustering) of representations for the entire dataset, which in turn increases resource requirements and may be intractable for web-scale data.

Online methods overcome this limitation by generating pseudo-labels (or cluster assignments) in-batch during training. 
These methods seamlessly scale to large datasets, but suffer the risk of samples being assigned to only a few (or even one) clusters, in which case they collapse to trivial solutions and fail to learn meaningful representations.
To prevent collapse, online clustering-based frameworks promote balanced clusters by applying appropriate balancing constraints on cluster assignments with methods such as centering~\cite{dino}, optimization~\cite{swav,mira} or auxiliary training objectives~\cite{twist}. 
Crucially,~\cite{twist,mira,swav} rely on in-batch statistics; that is, for each batch, only its own assignments are considered in applying the respective method's constraint.
As a result, these methods are highly reliant on large batch sizes, otherwise they face poor performance and training instability.
Furthermore,~\cite{dino,twist,mira} rely on soft cluster assignment statistics (e.g. the aggregate assignment confidence for each cluster) to define their constraints. This undermines the effectiveness of the constraints, as these metrics are only proxies for the actual size of the clusters.
We note that there are also hybrid methods such as~\cite{coke,smog}. These methods regulate cluster sizes in an online manner, but, for optimal performance, they both require storing and utilizing representations~\cite{smog} or cluster assignments~\cite{coke} for the entire dataset for each epoch, which again raises scalability issues with larger datasets.

In contrast to previous works, \ours combines the best of both worlds: it is fully online, and therefore scalable to any dataset size, \textit{and} it does not require large batch sizes. The latter, which is achieved by measuring cluster sizes over multiple batches explicitly with hard assignments, results in more accurate and reliable approximations of the cluster distribution across the dataset and improved training stability, even with a very small batch size.

\section{Method}\label{sec:method}

\begin{figure}[t]
\begin{center}
\includegraphics[width=0.8\linewidth]{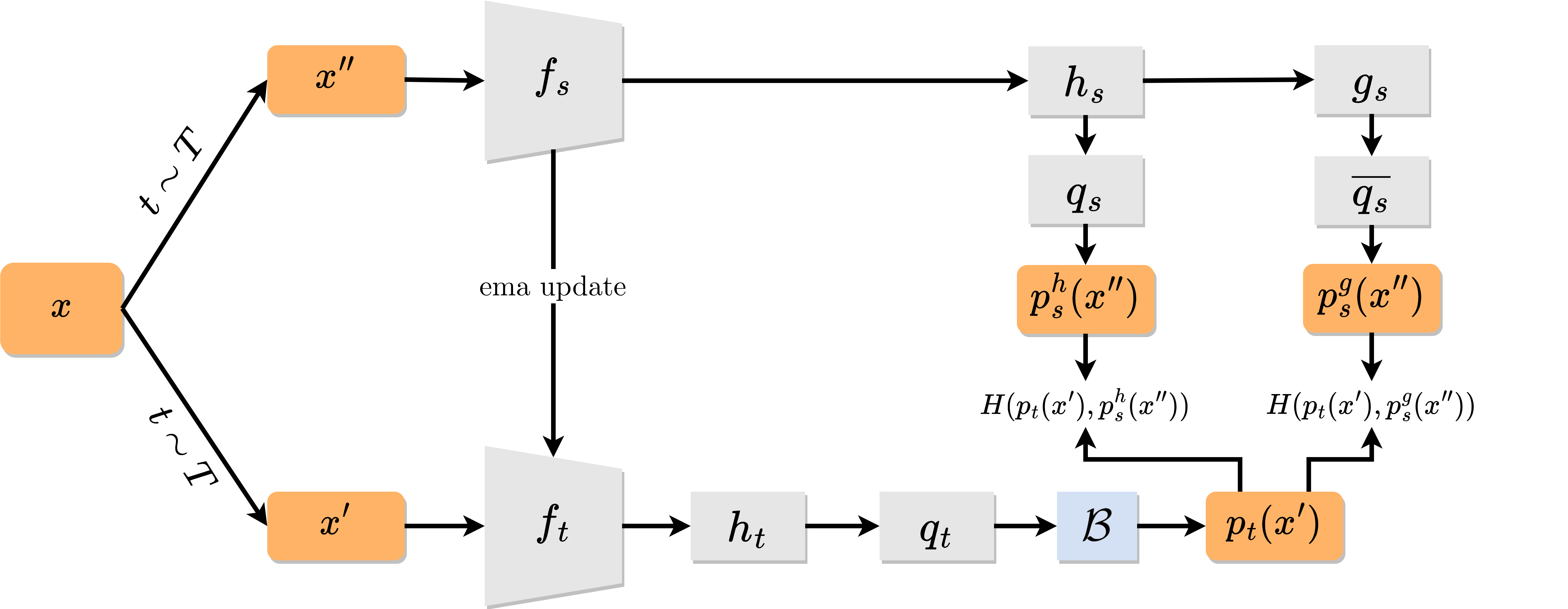}
\caption{Overview of \ours. The student is trained so that $\vp_s(x'')$ matches $\vp_t(x')$, where $x'$ and $x''$ are transformed views of $x$. The balancing module $\mathcal{B}$ adjusts cluster assignments to promote uniform distribution between the clusters \textit{across the dataset}.}
\label{fig:overview}
\end{center}
\end{figure}

We present an overview of \ours's architecture and objective in~\cref{sec:overview}. We then present our proposed, novel module for balancing cluster assignments in~\cref{sec:balancing}.

\subsection{Overview}\label{sec:overview}

\ours utilizes a teacher-student framework, where the student is trained to match the cluster assignments of the teacher. The student's architecture consists of a backbone model $f_s$, a projection MLP head $h_s$, a prediction MLP head $g_s$, and a layer $q_s$, which measures the cosine similarity between an input vector $\vv$ and a set of learned cluster centroids $\mC_s=[\vc_1, ..., \vc_K]\in\mathbb{R}^{K\times D}$, where $K$ is the number of clusters and $D$ the output dimension of $h_s$ and $g_s$:

\begin{equation}
    q_s(\vv, \vc) = \frac{\vv\cdot \vc}{|\vv| |\vc|}
\end{equation}

For each sample $x$, we then define the cosine similarity of its projection/prediction features with each cluster centroid as:

\begin{equation}
    \vz_s^h(x)=(q_s\circ h_s\circ f_s)(x) \ \ \ , \ \ \ \vz_s^g(x)=(\overline{q_s}\circ g_s\circ h_s\circ f_s)(x), 
\label{eq:z_s}
\end{equation}

\noindent where $\vz_s^h(x)$, $\vz_s^g(x)\in\mathbb{R}^{K}$ and $\overline{q_s}$ indicates that we apply stop-gradients to the centroid layer (i.e. the predictor considers the centroids as fixed).
The teacher has the same architecture but without a predictor, so that:

\begin{equation}
    \vz_t(x)=(q_t\circ h_t\circ f_t)(x).
\label{eq:z_t}
\end{equation}

\noindent We further define the adjusted sample-cluster cosine similarity $\vz^B_t$:

\begin{equation}
    \vz^B_t(x)=\mathcal{B}(\vz_t(x), \vs),
\label{eq:zb_t}
\end{equation}

\noindent where $\mathcal{B}$ is \ours's proposed balancing operator and $\vs$ is a vector measuring relative cluster size. This operator, presented in~\cref{sec:balancing}, adjusts sample-cluster similarities to promote balanced clusters in terms of size and prevent collapse.

Finally, the outputs of the student and teacher are the probability assignment vectors $\vp_s^h$, $\vp_s^g$ and $\vp_t\in\mathbb{R}^{K}$, mapping sample $x$ to each cluster $k \in K$:

\begin{equation}
    \vp_s^h(x)^{(k)} = \frac{exp\left(z_s^h(x)^{(k)}/\tau_s\right)}{\sum_{i=1}^{K}{exp(z_s^h(x)^{(i)}/\tau_s)}} \ \ \ , \ \ \ \vp_s^g(x)^{(k)} = \frac{exp(z_s^g(x)^{(k)}/\tau_s)}{\sum_{i=1}^{K}{exp(z_s^g(x)^{(i)}/\tau_s)}},
\label{eq:p_s}
\end{equation}

\begin{equation}
    \vp_t(x)^{(k)} = \frac{exp((z^B_t(x)^{(k)})/\tau_t)}{\sum_{i=1}^{K}{exp((z^B_t(x)^{(i)})/\tau_t)}},
\end{equation}

\noindent with $\tau_s$ and $\tau_t$ being temperature hyperparameters.

Following previous works~\cite{dino,mocov2,simsiam,mira}, the teacher's weights $\theta_t$ are updated in every training step following an exponential moving average (EMA) of the trained student's weights $\theta_s$, with update rule $\theta_t=m\theta_t+(1-m)\theta_s$.
The training objective of \ours~is then formally defined as maximizing the agreement between the student's and teacher's outputs across views, by minimizing the loss $L$:

\begin{equation}
\begin{aligned}
    L=\frac{1}{2}\sum_{x'\in G}{\ \sum_{\substack{x''\in G\cup L \\ x''\neq x'}}{H(\vp_t(x'), \vp^h_s(x''))}}+
    \frac{1}{2}\sum_{x'\in G}{\ \sum_{\substack{x''\in G\cup L}}{H(\vp_t(x'), \vp^g_s(x''))}}
\end{aligned},
\end{equation}
where $H(a,b)=-a^T~log~b$ and $G$, $L$ represent global and local views produced by random transformations $t\sim T$ applied to the original image $x$~\cite{swav}.

\subsection{Online Cluster Balancing}\label{sec:balancing}

The balancing operator $\mathcal{B}$ is applied to the teacher, as seen in~\cref{eq:zb_t}, in order to adjust sample-cluster similarities and promote clusters of equal size. Balancing operators are critical for online clustering-based self-supervised frameworks, as without it they are prone to collapsing to trivial solutions (i.e. the teacher assigning all samples to the same cluster)~\cite{swav,dino,mira}. 
In \ours, $\mathcal{B}$ consists of two components: a) keeping track of cluster assignments over multiple batches to estimate the clusters' relative size, and b) adjusting sample-cluster similarity $z$ accordingly, in order to regulate cluster assignments, and thereby promote balanced clusters and avoid collapse.

\paragraph{Measuring relative cluster sizes.} 

We define the relative cluster size vector $\vs\in \mathbb{R}^K$.
For each batch of $N_B$ samples $X$ we obtain the teacher's cluster assignments $\mP_t(X)=[\vp_t(x_1), ..., \vp_t(x_{N_B})]\in \mathbb{R}^{N_B\times K}$, and calculate the in-batch relative cluster size vector $\vs_B\in \mathbb{R}^{K}$ as the proportion of samples assigned to each cluster:

\begin{equation}\label{eq:defsx}
    \vs_B^{(k)}=\frac{1}{N_B}\sum_{n=1}^{N_B}{\1_\mathrm{\underset{k\in K}{argmax}(\vp_t(x_n))=k}} .
\end{equation}

\noindent The vector $\vs$ is then updated for each batch as:

\begin{equation}\label{eq:defs}
    \vs = \vs m_s + \vs_B (1-m_s),
\end{equation}
where $m_s$ is a momentum hyperparameter.
Essentially, $\vs\in [0,1]$ measures the proportion of samples assigned to each cluster over multiple batches with an exponential moving average whose window length is determined by $m_s$. This approach yields an accurate estimate of cluster sizes across the dataset, without requiring a large batch size. If samples are distributed among clusters with absolute uniformity, then $\vs^{(k)}\rightarrow \frac{1}{K}$ $\forall k \in K$, whereas $\vs^{(k)}<\frac{1}{K}$ for undersized clusters and $\vs^{(k)}>\frac{1}{K}$ for oversized clusters.

Our approach improves on previous works in two key respects. 
Firstly, we measure cluster sizes over multiple batches, as opposed to using in-batch statistics.
Secondly, we explicitly measure cluster sizes with hard assignments, rather than implicitly, through proxy metrics (e.g. aggregate sample-cluster similarity~\cite{dino}, assignment confidence~\cite{twist}).
This way, \ours measures cluster sizes \textbf{a)} more accurately, as measurements across batches are closer to the dataset-wide cluster distribution, \textbf{b)} with lower resource requirements, as a large batch size is not necessary for an accurate estimate, and \textbf{c)} more reliably, as using hard cluster assignments means the measurement is independent of factors such as the assignments' confidence, which fluctuates during training.

\paragraph{Adjusting sample-cluster similarities.}
The cluster assignment tracking vector $\vs$ provides an estimate of relative cluster sizes at any given step. We then use it to adjust cluster assignments so that the teacher will assign more (fewer) samples to undersized (oversized) clusters. To that end, we define the operator $\mathcal{B}$:

\begin{equation}\label{eq:b}
    z^B=\mathcal{B}(z ; s) =
    \begin{cases}
    1 - \left[1-z\right]s K \ \ , \ \ \text{if} \ \  s<\frac{1}{K} \\
    \left[1+z\right]\frac{1}{s K}-1 \ \ , \ \  \text{if} \ \ s>\frac{1}{K} \\
    z \ \  , \ \  \text{otherwise}
    \end{cases}
\end{equation}

\noindent where $z=\vz_t(x)^{(k)}$ and $s=\vs^{(k)}$ for a given cluster $k\in K$.

For any cluster $k$, $\mathcal{B}$ increases sample-cluster similarity if $k$ is undersized ($z^B>z$ for $s<\frac{1}{K}$), and decreases it if $k$ is oversized ($z^B<z$ for $s>\frac{1}{K}$). 
In this way, undersized (oversized) clusters are assigned more (fewer) samples, in an effort to approximate evenly sized clusters ($\vs^{(k)}\rightarrow1, \forall k\in K$).
We expand on the proposed balancing operator and the intuition behind it in~\cref{sec:bo}.

\paragraph{Summary.} 
Combined, the two components formulate a soft cluster balancing constraint that: a) prevents imbalanced cluster assignments that may lead to collapse and b) does not require a large batch size to be effective, as relative cluster sizes are measured over multiple batches. These advantages are validated through extensive experiments in~\cref{sec:experiments}, where we demonstrate that~\ours achieves state-of-the-art performance in representation learning benchmarks and is stable even when training with very small batch sizes.

\section{Experiments}\label{sec:experiments}

\subsection{Implementation details}

\paragraph{Architecture \& Hyperparameters.} 
Following~\cite{dino}, $h_s$ and $h_t$ are 2-layer MLPs with hidden dimension 2,048 and output dimension 256. $g_s$ is a 1-layer MLP with the same hidden/output dimensions.
The temperature parameters $\tau_t$ and $\tau_s$ are set to 0.04 and 0.1 respectively.
The teacher's update momentum $m$ is 0.996 and follows a cosine schedule to 1, and the momentum parameter $m_s$ of~\cref{eq:defs} is set to 0.999. 
When training without multi-crop, we extract two 224x224 views with scale range [0.14, 1.]. For multi-crop training, we extract two global 224x224 and six local 96x96 views with scale ranges [0.2, 1.] and [0.05, 0.2] respectively.
Unless stated otherwise, we set the number of clusters $K$ to 65,536, the batch size to 1,024 and use the same image augmentations as~\cite{byol}.
In our experiments, we primarily use a ResNet50 backbone, which is most frequently used in this task.
We further conduct experiments with a ViT-S/16~\cite{vit} backbone, to examine how \ours performs with transformer-based architectures.
When pretraining with a ResNet50 backbone, we use the SGD optimizer with momentum 0.9 and weight decay $10^{-4}$. The learning rate is set to 0.15$\times$batch size/256, and is linearly scaled for 10 epochs followed by a cosine decay schedule.
When pretraining with ViT, given limited resources, we \textbf{do not tune training hyperparameters} and instead follow the exact configuration used in~\cite{dino}.
Experiments are conducted on 4 A100 GPUs.

\paragraph{Learning cluster centroids.} It is well established that local views do not produce reliable labels, which is why target assignments are only produced by global views~\cite{swav}.
We extend this approach and only update cluster centroids through global views. This is equivalent to stop-gradient being applied to the centroid layer $q_s$ when processing local crops. We find that this leads to more stable training and, ultimately, better downstream performance. Furthermore, for ResNet50 pretraining, we use a distinct weight decay for the centroid layer $q_s$, following a cosine decay schedule from $10^{-3}$ to $10^{-4}$. The positive impact of these choices is examined in~\cref{tab:ablation_components}, as well as our use of a predictor, similar to~\cite{coke}.

\begin{table}[t]
\centering
\caption{\textbf{Linear \& k-NN classification on ImageNet.} We report linear and k-NN classification accuracy on ImageNet, along with each method's pretraining batch size and epochs. *TWIST follows standard pretraining with filtered self-labelled training.}
\label{tab:main}
\begin{adjustbox}{width=0.8\textwidth}
\begin{tabular}{l c c c c }
\toprule
\textbf{Method} & \textbf{Batch Size} & \textbf{Epochs} & \textbf{Linear} & \textbf{k-NN} \\
\midrule
Supervised & - & - & 75.6 & - \\
\arrayrulecolor{gray!30}\hline\arrayrulecolor{black}
SimSiam~\cite{simsiam} & 256 & 800 & 71.3 & - \\
SimCLR~\cite{simclr} & 4096 & 800 & 71.7 & - \\
BYOL~\cite{byol} & 4096 & 1000 & 74.4 & 64.8 \\
MoCo-v3~\cite{chen2021empirical} & 4096 & 1000 & 74.6 & - \\
DeepCluster v2~\cite{swav} & 4096 & 800 & 75.2 & - \\
Barlow Twins~\cite{barlowtwins} & 2048 & 1000 & 73.2 & 66.0 \\
VICReg~\cite{vicreg} & 2048 & 1000 & 73.2 & - \\
SwAV~\cite{swav} & 4096 & 800 & 75.3 & 65.7 \\
DINO~\cite{dino} & 4096 & 800 & 75.3 & 67.5 \\
NNCLR~\cite{nnclr} & 4096 & 1000 & 75.4 & \\
TWIST*~\cite{twist} & 2048 & 800+50 & 75.5 & - \\
MIRA~\cite{mira} & 4096 & 800 & 75.7 & 68.8 \\
MAST~\cite{huang2023mast} & 2048 & 1000 & 75.8 & - \\
CoKe~\cite{coke} & 1024 & 800 & \underline{76.4} & - \\
SMoG~\cite{smog} & 4096 & 400 & \underline{76.4} & - \\
\rowcolor{lime}
\textbf{\ours} & 1024 & 400 & \textbf{76.5} & \textbf{71.0} \\
\bottomrule
\end{tabular}
\end{adjustbox}
\end{table}

\subsection{Results}

In this section, we present results across standard evaluation benchmarks for self-supervised representation learning. Unless stated otherwise \ours is pretrained on ImageNet's train set for 400 epochs with multi-crop and batch size 1,024. 
In all cases, the best results are presented in \textbf{bold} and the second best are \underline{underlined}. Results for other methods are taken from~\cite{smog,mira,twist,huang2023mast,coke}.

\paragraph{Linear \& k-NN Classifier.}
We train a linear classifier on top of a frozen backbone on ImageNet's train set and present results on the validation set. For ResNet we follow the standard protocol~\cite{mocov2} and train the classifier for 100 epochs with the SGD optimizer, batch size 256 and learning rate 0.3 with cosine decay. We also present results for a k-NN classifier, where, following~\cite{dino}, we set the number of neighbours to 20. For the ViT backbone, we again follow~\cite{dino} and train for 100 epochs, with batch size 1,024, the SGD optimizer, and sweep the learning rate.

\setlength{\tabcolsep}{17pt}

\begin{table}[t]
\begin{center}
\caption{\textbf{Linear classification with ViT.} We report linear classification accuracy on ImageNet for various epochs.}
\label{tab:vit}
\begin{adjustbox}{width=0.7\textwidth}
\begin{tabular}{l c c c c }
\toprule
\multirow{2}{*}{\textbf{Method}} & \multirow{2}{*}{\textbf{Batch Size}} & \multicolumn{3}{c}{\textbf{Epochs}} \\
\cline{3-5}
 & & 100 & 300 & 800 \\
\midrule
MoCo-v3~\cite{chen2021empirical} & 4096 & - & 72.5 & - \\
DINO~\cite{dino} & 1024 & 73.8 & 75.9 & 77.0 \\
TWIST~\cite{twist} & 1024 & - & \underline{76.3} & - \\
\rowcolor{lime}
\textbf{\ours} & 1024 & \textbf{73.9} & \textbf{76.4} & \textbf{77.1} \\
\bottomrule
\end{tabular}
\end{adjustbox}
\end{center}
\end{table}

\setlength{\tabcolsep}{10pt}

\definecolor{gainsboro}{rgb}{0.86, 0.86, 0.86}

\begin{table}[t]
\begin{center}
\caption{\textbf{Linear classification on ImageNet for various settings.} We report linear classification accuracy on ImageNet for various epochs and with/without multi-crop.}
\label{tab:comprehensive}
\begin{adjustbox}{width=0.60\textwidth}
\begin{tabular}{l c c c c }
\toprule
\multirow{2}{*}{\textbf{Method}} &\multirow{2}{*}{\textbf{Batch Size}} & \multicolumn{3}{c}{\textbf{Epochs}} \\
\cline{3-5}
 & & 100 & 200 & 400 \\
\midrule
\rowcolor{gainsboro}
\multicolumn{5}{c}{\textit{Without Multi-Crop}} \\
BYOL & 4096 & 66.5 & 70.6 & 73.2 \\
SwAV & 4096 & 66.5 & 69.1 & 70.7 \\
VicReg & 2048 & 68.6 & 70.2 & 72.3 \\
MIRA & 4096 & 69.4 & \underline{72.3} & 73.3 \\
MAST & 2048 & - & 70.9 & 73.5 \\
TWIST & 2048 & \underline{70.4} & 70.9 & 71.8 \\
SMoG & 2048 & 67.2 & - & \underline{73.6} \\
\rowcolor{lime}
\textbf{\ours} & 1024 & \textbf{70.7} & \textbf{72.7} & \textbf{73.9} \\
\rowcolor{gainsboro}
\multicolumn{5}{c}{\textit{With Multi-Crop}} \\
SwAV & 4096 & 72.1 & 73.9 & 74.6 \\
TWIST & 2048 & 72.9 & 73.7 & 74.4 \\
MIRA & 4096 & \underline{73.6} & \underline{74.9} & 75.6 \\
SMoG & 4096 & - & - & \underline{76.4} \\
\rowcolor{lime}
\textbf{\ours} & 1024 & \textbf{74.5} & \textbf{75.7} & \textbf{76.5} \\
\bottomrule
\end{tabular}
\end{adjustbox}
\end{center}
\end{table}

The main results for this setting are presented in~\cref{tab:main,tab:vit} for ResNet50 and ViT backbones respectively. Additional, more extensive, comparisons, including results without multi-crop and for various pretraining epochs, are presented in~\cref{tab:comprehensive} for ResNet50.
Across settings,~\ours consistently outperforms previous works, achieving state-of-the-art performance. Importantly, this holds across pretraining epochs, as shown in~\cref{tab:comprehensive}. Furthermore, \ours outperforms previous works with a ViT backbone, despite using the hyperparameters suggested by~\cite{dino} without hyperparameter tuning. These results highlight \ours's efficiency in terms of training time, its effectiveness across architectures, and its robustness with regard to batch size and training hyperparameters.

\paragraph{Semi-supervised learning.}
Having evaluated \ours in a frozen backbone setting, we now present results for semi-supervised fine-tuning in~\cref{tab:semisup}, where a linear classifier and the backbone are trained on ImageNet's train set with limited labels. Following previous works, we train with 1\% and 10\% of labels using the splits specified in~\cite{simclr}, and report top-1 and top-5 accuracy on ImageNet's validation set. 
We fine-tune for 50 epochs with batch size 512, and use a backbone learning rate of 0.00008 and 0.0003 and a classification head learning rate of 1. and 0.2 for the 1\% and 10\% settings respectively.
In this setting as well, \ours achieves state-of-the-art results. We stress that this is achieved in a much more resource-efficient way relative to other works, and highlight that the two most competitive methods in~\cref{tab:semisup} are pretrained with a larger batch size~\cite{smog} and for more epochs~\cite{huang2023mast}.

\setlength{\tabcolsep}{6pt}

\begin{table}[t]
\begin{center}
\caption{\textbf{Semi-supervised finetuning on ImageNet.} We finetune on ImageNet with 1\% and 10\% of labels and report top-1 and top-5 accuracy on the validation set.}
\label{tab:semisup}
\resizebox{0.85\columnwidth}{!}{
\begin{tabular}{l c c c c c c }
\toprule
\multirow{2}{*}{\textbf{Method}} & \multirow{2}{*}{\textbf{Batch Size}} & \multirow{2}{*}{\textbf{Epochs}} & \multicolumn{2}{c}{\textbf{1\% Labels}} & \multicolumn{2}{c}{\textbf{10\% Labels}} \\
  &  &  & Top-1 & Top-5 & Top-1 & Top-5 \\
\midrule
Supervised & - & - & 25.4 & 48.4 & 56.4 & 80.4 \\
\arrayrulecolor{gray!30}\hline\arrayrulecolor{black}
BYOL~\cite{byol} & 4096 & 1000 & 53.2 & 78.4 & 68.8 & 89.0 \\
SwAV~\cite{swav} & 4096 & 800 & 53.9 & 78.5 & 70.2 & 89.9 \\
Barlow Twins~\cite{barlowtwins} & 2048 & 1000 & 55.0 & 79.2 & 69.7 & 89.3 \\
DINO~\cite{dino} & 4096 & 800 & 52.2 & 78.2 & 68.2 & 89.1 \\
NNCLR~\cite{nnclr} & 4096 & 1000 & 56.4 & 80.7 & 69.8 & 89.3 \\
MIRA~\cite{mira} & 4096 & 400 & 55.6 & 80.5 & 69.9 & 90.0 \\
MAST~\cite{huang2023mast} & 2048 & 1000 & 55.8 & 81.0 & \underline{71.4} & \textbf{90.9} \\
SMoG~\cite{smog} & 4096 & 400 & \textbf{58.0} & \underline{81.6} & 71.2 & 90.5 \\
\rowcolor{lime}
\textbf{\ours} & 1024 & 400 & \underline{57.8} & \textbf{81.8} & \textbf{71.5} & \underline{90.7} \\
\bottomrule
\end{tabular}
}
\end{center}
\end{table}

\setlength{\tabcolsep}{4pt}

\begin{table}[t]
\begin{center}
\caption{\textbf{Object Detection \& Segmentation.} We use a ResNet50 backbone pretrained on ImageNet to initialize a Mask R-CNN~\cite{maskrcnn} detector. We train it on MS COCO {\tt train2017} and present results on {\tt val2017}.}
\label{tab:detection}
\resizebox{0.85\columnwidth}{!}{
\begin{tabular}{l c c c c c c c c}
\toprule
\textbf{Method} & \textbf{Batch Size} & \textbf{Epochs} & $AP^b$ & $AP^b_{50}$ & $AP^b_{75}$ & $AP^m$ & $AP^m_{50}$ & $AP^m_{75}$ \\
\midrule
Supervised & - & - & 38.2 & 58.2 & 41.2 & 33.3 & 54.7 & 35.2 \\
\arrayrulecolor{gray!30}\hline\arrayrulecolor{black}
MoCo-v2~\cite{mocov2} & 256 & 800 & \textbf{39.3} & 58.9 & \textbf{42.5} & \textbf{34.4} & 55.8 & \underline{36.5} \\
SwAV~\cite{swav} & 4096 & 800 & 38.4 & 58.6 & 41.3 & 33.8 & 55.2 & 35.9 \\
DINO~\cite{dino} & 4096 & 800 & 37.4 & 57.8 & 40.0 & 33.0 & 54.3 & 34.9 \\
SimSiam~\cite{simsiam} & 256 & 800 & \underline{39.2} & \underline{59.3} & 42.1 & \textbf{34.4} & \underline{56.0} & \textbf{36.7} \\
BarlowTwins~\cite{barlowtwins} & 2048 & 1000 & 39.2 & 59.0 & \textbf{42.5} & 34.3 & \underline{56.0} & \underline{36.5} \\
TWIST*~\cite{twist} & 2048 & 800+50 & 38.0 & 58.4 & 40.8 & 33.5 & 54.9 & 35.5 \\
\rowcolor{lime}
\textbf{\ours} & 1024 & 400 & \underline{39.2} & \textbf{59.4} & 42.3 & 34.3 & \textbf{56.2} & 36.3 \\
\bottomrule
\end{tabular}
}
\end{center}
\end{table}

\paragraph{Object Detection \& Instance Segmentation.}
Finally, we evaluate \ours on dense prediction tasks, specifically object detection and instance segmentation. Following previous works, we train Mask R-CNN~\cite{maskrcnn} with a C4 backbone on MS COCO~\cite{coco2014} {\tt train2017} for the standard $1\times$ schedule and present results for detection ($AP^b$) and segmentation ($AP^m$) on MS COCO {\tt val2017} in~\cref{tab:detection}.
We find that \ours performs competitively with previous works, achieving best or second best outcomes for most metrics, and note that \ours achieves this performance while having been trained for much fewer epochs than other methods.

\setlength{\tabcolsep}{10pt}

\subsection{Ablations \& Analysis}

In this section, we analyze the impact of \ours's key components and its training properties. Unless stated otherwise, \ours is pretrained with ResNet50, for 100 epochs and without multi-crop.

\paragraph{Batch size.}
A key advantage of \ours over previous methods is that it achieves state-of-the-art results with a much smaller batch size (1,024) than is typically used for pretraining. To further examine this property, we train \ours with even smaller batch sizes, and present results in~\cref{tab:batch_size}. We observe that, even for a batch size of 256 and without any hyperparameter tuning, \ours suffers minimal performance drop, maintaining state-of-the-art performance for 100 epochs pretraining without multi-crop. This is a strong indication that \ours is indeed remarkably effective and stable with small batch training.

\paragraph{Number of clusters.}
We ablate the number of clusters $K$ used to pretrain \ours and present results in~\cref{tab:cluster_no}. We observe that the optimal number is $K=65,536$, similarly with~\cite{dino}, with minor performance drops for smaller and larger $K$ values.

\setlength{\tabcolsep}{23pt}

\begin{table}[t]
\begin{center}
\caption{Linear evaluation accuracy on Imagenet set for varying batch sizes.}
\label{tab:batch_size}
\resizebox{0.7\textwidth}{!}{
\begin{tabular}{l c c c}
\toprule
\textbf{Batch Size} & 256 & 512 & 1024 \\ 
\midrule
\textbf{Linear Acc.} & 70.5 & 70.6 & \textbf{70.7} \\
\bottomrule
\end{tabular}
}
\end{center}
\end{table}

\setlength{\tabcolsep}{10pt}

\setlength{\tabcolsep}{18pt}

\begin{table}[t]
\begin{center}
\caption{Linear evaluation accuracy on Imagenet for varying cluster numbers $K$.}
\label{tab:cluster_no}
\resizebox{0.75\columnwidth}{!}{
\begin{tabular}{l c c c c }
\toprule
\textbf{Number of clusters K} & 32768 & 49152 & 65536 & 98304 \\ 
\midrule
\textbf{Linear Acc.} & 70.3 & 70.5 & \textbf{70.7} & 70.6 \\
\bottomrule
\end{tabular}
}
\end{center}
\end{table}

\setlength{\tabcolsep}{10pt}

\setlength{\tabcolsep}{20pt}

\begin{table}[b]
\begin{center}
\caption{Linear evaluation accuracy on Imagenet for~\ours \textit{with} multi-crop, ablating on various components.}
\label{tab:ablation_components}
\resizebox{0.95\columnwidth}{!}{
\begin{tabular}{l c c c}
\toprule
\textbf{Predictor} & \textbf{q$_s$ Stop-gradient} & \textbf{q$_s$ WD} & \textbf{Linear Acc.} \\ 
\midrule
\xmark & \xmark & $0.0001$ & 73.8 \\
\cmark & \xmark & $0.0001$ & 74.0 \\
\cmark & \cmark & $0.0001$ & 74.3 \\
\cmark & \cmark & $0.001\rightarrow 0.0001$ & \textbf{74.5} \\
\bottomrule
\end{tabular}
}
\end{center}
\end{table}

\setlength{\tabcolsep}{10pt}

\paragraph{Architecture \& hyperparameters.}
In~\cref{tab:ablation_components} we present the impact of individual components of \ours. Specifically, we examine the impact of including a predictor, not updating $q_s$ for local crops, and having a separate weight decay schedule for $q_s$. We observe that each component in turn improves performance, even though, notably, \ours's baseline performance is already high and competitive with other works in the 100 pretraining epochs setting (see~\cref{tab:comprehensive}).

\begin{figure}
    \centering
    
    \begin{subfigure}{0.32\linewidth}
    \centering
    \begin{tikzpicture}
        \begin{axis}[
            xmin=0, xmax=410,
            ymin=4.5, ymax=11.,
            xtick={0, 200, 400},
            ytick={4, 6, 8, 10},
            axis lines=left,
            width=1.\linewidth,
        ]

        \draw[dashed, gray!60, line width=0.75pt] (axis cs:100,\pgfkeysvalueof{/pgfplots/ymin}) -- (axis cs:100,\pgfkeysvalueof{/pgfplots/ymax});
        \draw[dashed, gray!60, line width=0.75pt] (axis cs:200,\pgfkeysvalueof{/pgfplots/ymin}) -- (axis cs:200,\pgfkeysvalueof{/pgfplots/ymax});
        \draw[dashed, gray!60, line width=0.75pt] (axis cs:300,\pgfkeysvalueof{/pgfplots/ymin}) -- (axis cs:300,\pgfkeysvalueof{/pgfplots/ymax});
        \draw[dashed, gray!60, line width=0.75pt] (axis cs:400,\pgfkeysvalueof{/pgfplots/ymin}) -- (axis cs:400,\pgfkeysvalueof{/pgfplots/ymax});

        \draw[dashed, gray!60, line width=0.75pt] (axis cs:\pgfkeysvalueof{/pgfplots/xmin}, 4) -- (axis cs:\pgfkeysvalueof{/pgfplots/xmax}, 4);
        \draw[dashed, gray!60, line width=0.75pt] (axis cs:\pgfkeysvalueof{/pgfplots/xmin}, 6) -- (axis cs:\pgfkeysvalueof{/pgfplots/xmax}, 6);
         \draw[dashed, gray!60, line width=0.75pt] (axis cs:\pgfkeysvalueof{/pgfplots/xmin}, 8) -- (axis cs:\pgfkeysvalueof{/pgfplots/xmax}, 8);
         \draw[dashed, gray!60, line width=0.75pt] (axis cs:\pgfkeysvalueof{/pgfplots/xmin}, 10) -- (axis cs:\pgfkeysvalueof{/pgfplots/xmax}, 10);
                 
        \addplot[blue, smooth, line width=0.75pt] table [y index=1, x expr=\coordindex, col sep=comma, skip first n=1] {figures/statistics.csv};
        
        \end{axis}
    \end{tikzpicture}
    \captionsetup{justification=centering} 
    \caption{Training \\ \hspace{10pt} loss}\label{fig:loss}
    \end{subfigure}
    \hfill
    \begin{subfigure}{0.32\linewidth}
    \centering
    \begin{tikzpicture}
        \begin{axis}[
            xmin=0, xmax=410,
            ymin=0, ymax=0.3,
            xtick={0, 200, 400},
            ytick={0.05, 0.15, 0.25},
            yticklabels={0.05, 0.15, 0.25},
            axis lines=left,
            width=1.\linewidth,
        ]

        \draw[dashed, gray!60, line width=0.75pt] (axis cs:100,\pgfkeysvalueof{/pgfplots/ymin}) -- (axis cs:100,\pgfkeysvalueof{/pgfplots/ymax});
        \draw[dashed, gray!60, line width=0.75pt] (axis cs:200,\pgfkeysvalueof{/pgfplots/ymin}) -- (axis cs:200,\pgfkeysvalueof{/pgfplots/ymax});
        \draw[dashed, gray!60, line width=0.75pt] (axis cs:300,\pgfkeysvalueof{/pgfplots/ymin}) -- (axis cs:300,\pgfkeysvalueof{/pgfplots/ymax});
        \draw[dashed, gray!60, line width=0.75pt] (axis cs:400,\pgfkeysvalueof{/pgfplots/ymin}) -- (axis cs:400,\pgfkeysvalueof{/pgfplots/ymax});

        \draw[dashed, gray!60, line width=0.75pt] (axis cs:\pgfkeysvalueof{/pgfplots/xmin}, 0.05) -- (axis cs:\pgfkeysvalueof{/pgfplots/xmax}, 0.05);
        \draw[dashed, gray!60, line width=0.75pt] (axis cs:\pgfkeysvalueof{/pgfplots/xmin}, 0.15) -- (axis cs:\pgfkeysvalueof{/pgfplots/xmax}, 0.15);
         \draw[dashed, gray!60, line width=0.75pt] (axis cs:\pgfkeysvalueof{/pgfplots/xmin}, 0.25) -- (axis cs:\pgfkeysvalueof{/pgfplots/xmax}, 0.25);

        \addplot[blue, smooth, line width=0.75pt] table [y index=4, x expr=\coordindex, col sep=comma, skip first n=1] {figures/statistics.csv};
        \addplot[blue, smooth, line width=0.75pt] table [y index=4, x expr=\coordindex, col sep=comma, skip first n=1] {figures/statistics.csv};
        
        \end{axis}
    \end{tikzpicture}
    \captionsetup{justification=centering}
    \caption{Teacher \\ \hspace{10pt} confidence}\label{fig:confidence}
    \end{subfigure}
    \hfill
    \begin{subfigure}{0.32\linewidth}
    \centering
    \begin{tikzpicture}
        \begin{axis}[
            xmin=0, xmax=410,
            ymin=0.5, ymax=1.5,
            xtick={0, 200, 400},
            ytick={0.7, 1., 1.3},
            axis lines=left,
            width=1.\linewidth,
        ]

        \draw[dashed, gray!60, line width=0.75pt] (axis cs:100,\pgfkeysvalueof{/pgfplots/ymin}) -- (axis cs:100,\pgfkeysvalueof{/pgfplots/ymax});
        \draw[dashed, gray!60, line width=0.75pt] (axis cs:200,\pgfkeysvalueof{/pgfplots/ymin}) -- (axis cs:200,\pgfkeysvalueof{/pgfplots/ymax});
        \draw[dashed, gray!60, line width=0.75pt] (axis cs:300,\pgfkeysvalueof{/pgfplots/ymin}) -- (axis cs:300,\pgfkeysvalueof{/pgfplots/ymax});
        \draw[dashed, gray!60, line width=0.75pt] (axis cs:400,\pgfkeysvalueof{/pgfplots/ymin}) -- (axis cs:400,\pgfkeysvalueof{/pgfplots/ymax});

        \draw[dashed, gray!60, line width=0.75pt] (axis cs:\pgfkeysvalueof{/pgfplots/xmin}, 0.7) -- (axis cs:\pgfkeysvalueof{/pgfplots/xmax}, 0.7);
        \draw[dashed, gray!60, line width=0.75pt] (axis cs:\pgfkeysvalueof{/pgfplots/xmin}, 1.) -- (axis cs:\pgfkeysvalueof{/pgfplots/xmax}, 1.);
         \draw[dashed, gray!60, line width=0.75pt] (axis cs:\pgfkeysvalueof{/pgfplots/xmin}, 1.3) -- (axis cs:\pgfkeysvalueof{/pgfplots/xmax}, 1.3);

        \addplot[blue, smooth, line width=0.75pt] table [y index=8, x expr=\coordindex, col sep=comma, skip first n=1] {figures/statistics.csv};
        \addplot[blue, smooth, line width=0.75pt] table [y index=9, x expr=\coordindex, col sep=comma, skip first n=1] {figures/statistics.csv};
        
        \end{axis}
    \end{tikzpicture}
    \captionsetup{justification=centering} 
    \caption{Min. and Max. relative cluster sizes}\label{fig:cluster_sizes}
    \end{subfigure}
    \vfill
    \hspace{1cm}
    \begin{subfigure}[t]{0.32\linewidth}
    \centering
    \begin{tikzpicture}
        \begin{axis}[
            xmin=0, xmax=410,
            ymin=0, ymax=0.55,
            xtick={0, 200, 400},
            ytick={0, 0.1, 0.3, 0.5},
            axis lines=left,
            width=1.\linewidth, 
        ]

        \draw[dashed, gray!60, line width=0.75pt] (axis cs:100,\pgfkeysvalueof{/pgfplots/ymin}) -- (axis cs:100,\pgfkeysvalueof{/pgfplots/ymax});
        \draw[dashed, gray!60, line width=0.75pt] (axis cs:200,\pgfkeysvalueof{/pgfplots/ymin}) -- (axis cs:200,\pgfkeysvalueof{/pgfplots/ymax});
        \draw[dashed, gray!60, line width=0.75pt] (axis cs:300,\pgfkeysvalueof{/pgfplots/ymin}) -- (axis cs:300,\pgfkeysvalueof{/pgfplots/ymax});
        \draw[dashed, gray!60, line width=0.75pt] (axis cs:400,\pgfkeysvalueof{/pgfplots/ymin}) -- (axis cs:400,\pgfkeysvalueof{/pgfplots/ymax});

        \draw[dashed, gray!60, line width=0.75pt] (axis cs:\pgfkeysvalueof{/pgfplots/xmin}, 0.1) -- (axis cs:\pgfkeysvalueof{/pgfplots/xmax}, 0.1);
        \draw[dashed, gray!60, line width=0.75pt] (axis cs:\pgfkeysvalueof{/pgfplots/xmin}, 0.3) -- (axis cs:\pgfkeysvalueof{/pgfplots/xmax}, 0.3);
        \draw[dashed, gray!60, line width=0.75pt] (axis cs:\pgfkeysvalueof{/pgfplots/xmin}, 0.5) -- (axis cs:\pgfkeysvalueof{/pgfplots/xmax}, 0.5);
                 
        \addplot[blue, smooth, line width=0.75pt] table [y index=5, x expr=\coordindex, col sep=comma, skip first n=1] {figures/statistics.csv};
        
        \end{axis}
    \end{tikzpicture}
    \captionsetup{justification=centering}
    \caption{Teacher assignments' agreement}\label{fig:aggreement}
    \end{subfigure}
    \hfill
    \begin{subfigure}[t]{0.32\linewidth}
    \centering
    \begin{tikzpicture}
        \begin{axis}[
            xmin=0, xmax=410,
            ymin=0., ymax=0.75,
            xtick={0, 200, 400},
            ytick={0., 0.3, 0.5, 0.7},
            axis lines=left,
            width=1.\linewidth, 
        ]

        \draw[dashed, gray!60, line width=0.75pt] (axis cs:100,\pgfkeysvalueof{/pgfplots/ymin}) -- (axis cs:100,\pgfkeysvalueof{/pgfplots/ymax});
        \draw[dashed, gray!60, line width=0.75pt] (axis cs:200,\pgfkeysvalueof{/pgfplots/ymin}) -- (axis cs:200,\pgfkeysvalueof{/pgfplots/ymax});
        \draw[dashed, gray!60, line width=0.75pt] (axis cs:300,\pgfkeysvalueof{/pgfplots/ymin}) -- (axis cs:300,\pgfkeysvalueof{/pgfplots/ymax});
        \draw[dashed, gray!60, line width=0.75pt] (axis cs:400,\pgfkeysvalueof{/pgfplots/ymin}) -- (axis cs:400,\pgfkeysvalueof{/pgfplots/ymax});

        \draw[dashed, gray!60, line width=0.75pt] (axis cs:\pgfkeysvalueof{/pgfplots/xmin}, 0.3) -- (axis cs:\pgfkeysvalueof{/pgfplots/xmax}, 0.3);
        \draw[dashed, gray!60, line width=0.75pt] (axis cs:\pgfkeysvalueof{/pgfplots/xmin}, 0.5) -- (axis cs:\pgfkeysvalueof{/pgfplots/xmax}, 0.5);
         \draw[dashed, gray!60, line width=0.75pt] (axis cs:\pgfkeysvalueof{/pgfplots/xmin}, 0.7) -- (axis cs:\pgfkeysvalueof{/pgfplots/xmax}, 0.7);

        \addplot[blue, smooth, line width=0.75pt] table [y index=6, x expr=\coordindex, col sep=comma, skip first n=1] {figures/statistics.csv};
        
        \end{axis}
    \end{tikzpicture}
    \captionsetup{justification=centering} 
    \caption{Teacher assignments' purity}\label{fig:accuracy}
    \end{subfigure}
        \hspace{1cm}
    \caption{\ours training statistics.}\label{fig:statistics}
\end{figure}
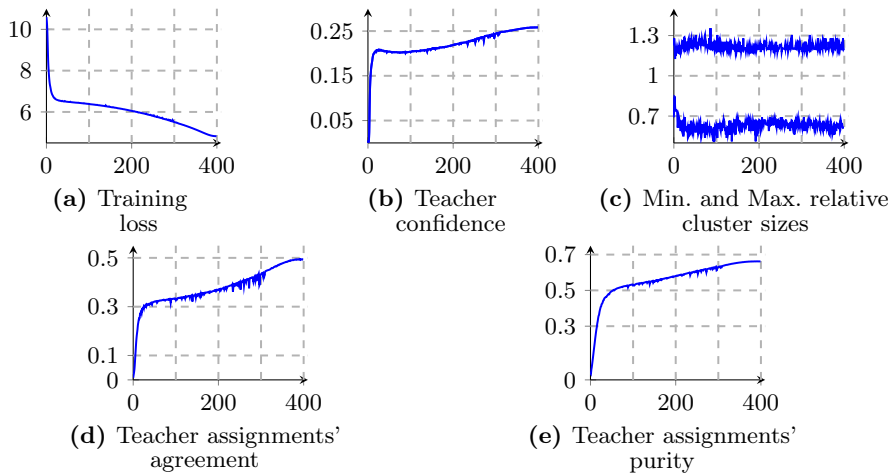

\paragraph{Training statistics.}
To provide insights into \ours and facilitate future research, we complement our experimental results with information about \ours's training. We analyze a 400-epoch training run with multi-crop, and present in~\cref{fig:statistics}: 
\noindent \textbf{a)} \ours's training loss (\cref{fig:loss}).
\noindent \textbf{b)} The cluster assignments' confidence $c_{avg}$ (\cref{fig:confidence}), measured as in~\cref{eq:confidence}.
\noindent \textbf{c)} The relative minimum/maximum cluster sizes for each epoch (\cref{fig:cluster_sizes}), measured as the ratio of the biggest/smallest cluster to the "optimal" size of $1/K$ samples. 
\noindent \textbf{d)} The average agreement of the teacher's assignments between views $a_{avg}$ (\cref{fig:aggreement}), measured as in~\cref{eq:aggreement}.
\noindent \textbf{e)} The purity~\cite{zhao2001criterion} of the clusters with regard to the ground truth labels (\cref{fig:accuracy}), which indicates how semantically meaningful they are.

\begin{equation}\label{eq:confidence}
    c_{avg}=\frac{1}{NG}\sum_{n=1}^{N}{\sum_{x_n'\in G}{{max(\vp_t(x_n'))}}}
\end{equation}
\begin{equation}\label{eq:aggreement}
    a_{avg}=\frac{1}{NG}\sum_{n=1}^{N}\sum_{x_n'\in G}\sum_{\substack{x_n''\in G \\ x_n''\neq x_n'}}{\1_\mathrm{\underset{k\in K}{argmax}(\vp_t(x_n'))=\underset{k\in K}{argmax}(\vp_t(x_n''))}}
\end{equation}

As seen in~\cref{fig:loss,fig:confidence}, \ours converges smoothly and without volatility, which highlights its stability. 
Furthermore, we observe in~\cref{fig:aggreement,fig:accuracy} that, during training, cluster assignments are increasingly reliable and semantically meaningful, as they become more consistent between views and increasingly include images with the same ground truth label.
Finally, \cref{fig:cluster_sizes} demonstrates the effectiveness of our balancing method. Throughout the training, clusters remain between 70\% and 130\% of the optimal size, which, for ImageNet and K=65,536, translates to clusters being assigned approximately between 13 and 25 samples.
The fact that \ours operates consistently within such narrow margins is a strong indicator of its effectiveness and stability in terms of cluster balancing.
Furthermore, we observe that, in practice, \ours softly enforces lower and upper bounds on cluster sizes rather than absolute uniformity. 
This is a positive property, as~\cite{coke} observed that enforcing balancing too strictly is detrimental to representation learning. 
Notably, in~\ours the upper/lower bounds emerge dynamically during training, as opposed to~\cite{coke}, where they are key hyperparameters that need to be defined by users.

\section{Discussion}

As was outlined in~\cref{sec:related_works,sec:method,sec:experiments}, \ours combines the best features of previous methods: \textbf{a)} it is online, therefore scalable to large datasets, \textbf{b)} it measures hard cluster assignments over multiple batches, which leads to reliable cluster size estimates even with small batch sizes, and \textbf{c)} it effectively balances clusters throughout training. 
This leads to state-of-the-art results in the primary, classification-related representation learning benchmarks, which include linear classification and semi-supervised learning. Additionally, \ours demonstrates competitive performance in dense prediction tasks (i.e. object detection and segmentation), despite training for fewer epochs compared to other works.
Furthermore, we apply \ours to pretraining a ViT backbone and obtain excellent performance, even though no hyperparameter tuning was applied.
These results indicate that \ours is highly effective and versatile, and can be reliably used for pretraining with different architectures and for various downstream tasks.

Beyond its performance, a key feature of \ours is its remarkable training efficiency in terms of the resources it requires. Regarding VRAM, our experiments demonstrate that \ours is stable and effective with very small batch sizes, achieving state-of-the-art results with a batch size of 1,024 -- by comparison, most other methods use a batch size 2X or 4X as large (with a corresponding increase in GPU memory utilization). With regard to training time, \ours again demonstrates remarkable efficiency, outperforming methods that were trained for at least 2X as many epochs. We attribute this to our novel balancing module $\mathcal{B}$: due to its more accurate and reliable estimation of cluster sizes and its smoother method of regulating them by adjusting their assignments, the model's supervision is much more stable throughout training. This facilitates convergence and results in better performance with less training time.

\section{Conclusion}

We present \ours, a novel clustering-based framework for self-supervised representation learning. \ours relies on a novel cluster balancing method that explicitly measures their sizes across multiple batches, and adjusts their assignments to promote evenly sized clusters. We conduct extensive experiments and find that \ours achieves state-of-the-art results across benchmarks and backbone architectures. However, crucially, our experiments demonstrate that \ours is also remarkably efficient, as it achieves the strong performance reported in this paper with less training and a much smaller batch size than most other frameworks. Overall, we believe that the proposed framework is not only significant in terms of its performance, but also as a step toward decreasing the resources required for self-supervised pretraining with visual data.
\\ \\
\noindent \textbf{Acknowledgements.}
This work was supported by the EU H2020 AI4Media No.951911 project. 
This research utilised Queen Mary's Apocrita HPC facility, supported by QMUL Research-IT. http://doi.org/10.5281/zenodo.438045.

\bibliographystyle{splncs04}
\bibliography{main}

\newpage

\appendix

\chapter*{Appendix}

The structure of the Appendix is as follows: 
In~\cref{sec:balancing_viz}, we study the effectiveness of \ours's cluster balancing module, by examining the relative cluster sizes \textit{after} training and comparing it with a similar work, DINO~\cite{dino}.
In~\cref{sec:bo} the balancing operator $\mathcal{B}$ is analyzed in depth, including the intuition behind our design choices and ablations regarding their effectiveness.
Subsequently, in~\cref{sec:embedding_viz}, we present a visualization of \ours's learned feature representations, demonstrating that classes are well separated despite the training process being unsupervised.
Finally, to facilitate understanding of \ours, we provide pseudo-code for a training step in Pytorch style in~\cref{sec:algorithm}.

\section{Cluster Balancing Effectiveness}\label{sec:balancing_viz}

\begin{figure}[t]
    \centering
    \begin{subfigure}{0.9\linewidth}
    \centering
    \includegraphics[width=1.\linewidth]{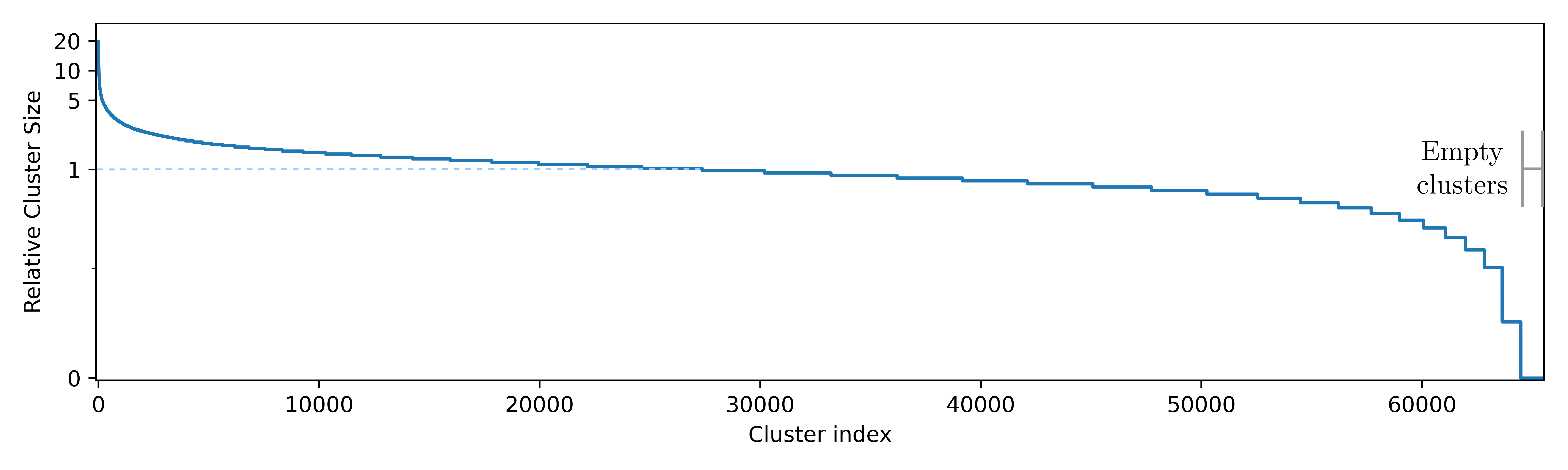}
    \captionsetup{justification=centering}
    \caption{Sample distribution for \ours. 1,056 clusters (1.6\%) are empty. The largest cluster has a relative size of 19.5 (i.e. is assigned 382 samples).}
    \label{fig:cd_mine}
    \end{subfigure}
    \vfill
    \begin{subfigure}{0.9\linewidth}
    \centering
    \includegraphics[width=1.\linewidth]{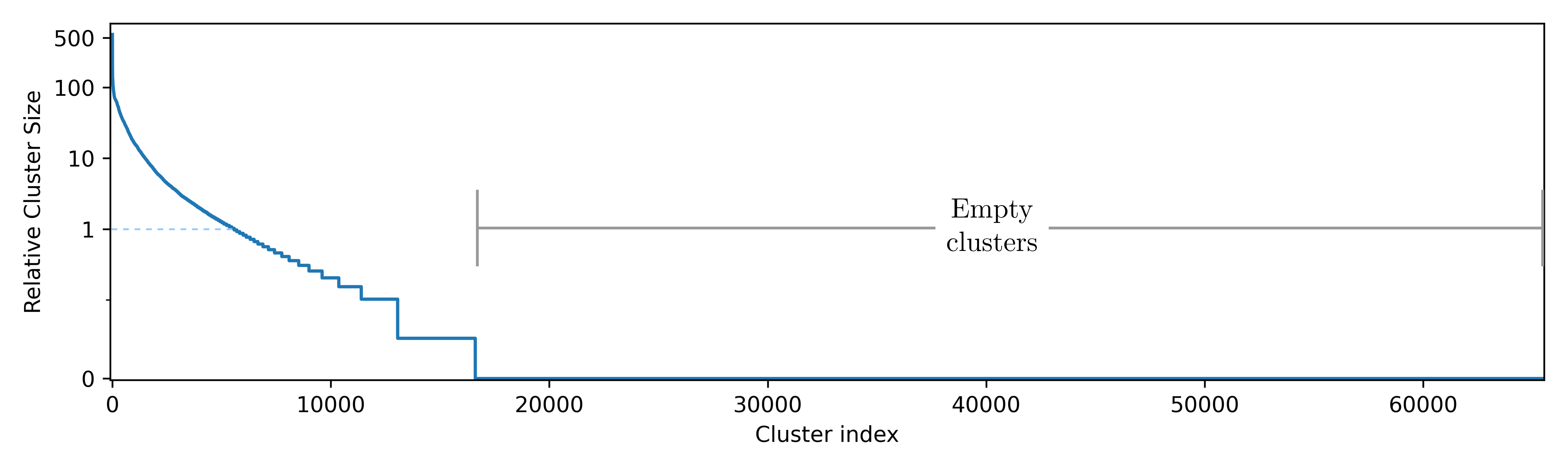}
    \captionsetup{justification=centering}
    \caption{Sample distribution for DINO. 48,925 clusters (74.7\%) are empty. The largest cluster has a relative size of 555.9 (i.e. is assigned 10,946 samples).}
    \label{fig:cd_dino}
    \end{subfigure}
    \caption{Sample distribution over the clusters for \ours and DINO. 
    Clusters are sorted according to their relative size, defined as $\frac{N_c K}{N}$, where $N_c$ is the number of samples assigned to that cluster, N=1,281,167 is the number of total samples and K=65,536 is the number of clusters. 
    In each plot, we highlight the optimal relative cluster size of 1 (for $N_c=\frac{N}{K}$) and the empty clusters ($N_c=0$).
    }
    \label{fig:cluster_distribution}
\end{figure}

In this section, we expand on the effectiveness of \ours's cluster balancing method. 
We present in~\cref{fig:cluster_distribution} the distribution of samples over the clusters for \ours and contrast it with DINO~\cite{dino}, a landmark work in the area of clustering-based self-supervised frameworks. 
Both \ours and DINO pretrain with $K=65,536$ clusters, which facilitates a fair comparison. 
Differently from the results presented in~\cref{sec:experiments}, where cluster measurements were made during training (i.e. over an epoch with the balancing module active), the measurements presented in~\cref{fig:cluster_distribution} were made via simple inference, using each model's final weights.
The assignments are calculated on ImageNet's train set. For DINO we use the publicly available weights provided for a ViT-S/16 backbone trained for 800 epochs with multi-crop.

As seen in~\cref{fig:cluster_distribution}, for \ours only 1.6\% of clusters are empty, contrasted with 74.7\% for DINO. 
Additionally, the samples among non-empty clusters are much more evenly distributed, as seen by contrasting~\cref{fig:cd_mine,fig:cd_dino}.
These findings complement the ones presented in~\cref{sec:experiments}, and reinforce our claims regarding the effectiveness of~\ours's cluster balancing approach, which we consider critical for the state-of-the-art performance and remarkable training efficiency of \ours.
Finally, we believe these findings highlight the effectiveness of our choice to balance clusters based on explicitly measuring their relative size, as opposed to relying on proxy metrics, such as, in the case of DINO, the "centerness" of sample-cluster similarities.

\section{Balancing Operator $\mathcal{B}$} \label{sec:bo}

The balancing operator $\mathcal{B}$ adjusts sample-cluster cosine similarities according to the relative cluster size vector $\vs$, as shown in the main paper's Eq. 10. 
Specifically, as the original sample-cluster similarity $z\in [-1,1]$, in Eq. 10 we first shift it to a positive range of $[0,2]$, adjust it depending on whether $s>\frac{1}{K}$ or $s<\frac{1}{K}$, and then shift it back to its original range of $[-1,1]$ to obtain the final similarity value $z^B$.

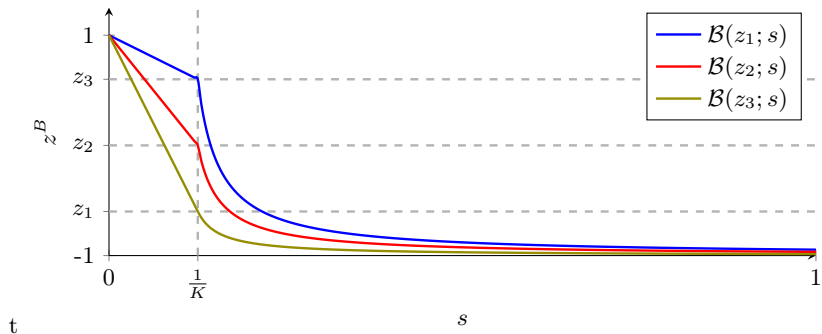
\begin{figure}{t}
    \centering
    \begin{tikzpicture}
        \begin{axis}[
            xlabel={$s$},
            ylabel={$z^B$},
            xmin=0, xmax=199,
            ymin=-1, ymax=1.25,
            xtick={0, 25, 199},
            xticklabels={0, $\frac{1}{K}$, 1},
            ytick={-1, -0.6, 0, 0.6, 1},
            yticklabels={-1, $z_1$, $z_2$, $z_3$, 1},
            axis lines=left,
            width=0.9\textwidth,
            height=0.4\textwidth,
        ]
        \draw[dashed, gray!60, line width=0.9pt] (axis cs:25, \pgfkeysvalueof{/pgfplots/ymin}) -- (axis cs:25, \pgfkeysvalueof{/pgfplots/ymax});
        \draw[dashed, gray!60, line width=0.9pt] (axis cs:\pgfkeysvalueof{/pgfplots/xmin}, 0) -- (axis cs:\pgfkeysvalueof{/pgfplots/xmax}, 0);
        \draw[dashed, gray!60, line width=0.9pt] (axis cs:\pgfkeysvalueof{/pgfplots/xmin}, 0.6) -- (axis cs:\pgfkeysvalueof{/pgfplots/xmax}, 0.6);
        \draw[dashed, gray!60, line width=0.9pt] (axis cs:\pgfkeysvalueof{/pgfplots/xmin}, -0.6) -- (axis cs:\pgfkeysvalueof{/pgfplots/xmax}, -0.6);
        
        \addplot[blue, smooth, line width=0.9pt] table [y index=0, x expr=\coordindex, col sep=comma] {figures/b_z_values.csv};
        \addlegendentry{$\mathcal{B}(z_1;s)$}

        \addplot[red, smooth, line width=0.9pt] table [y index=1, x expr=\coordindex, col sep=comma] {figures/b_z_values.csv};
        \addlegendentry{$\mathcal{B}(z_2;s)$}

        \addplot[olive, smooth, line width=0.9pt] table [y index=2, x expr=\coordindex, col sep=comma] {figures/b_z_values.csv};
        \addlegendentry{$\mathcal{B}(z_3;s)$}
        
        \end{axis}
    \end{tikzpicture}
    \caption{The values of $z^B=\mathcal{B}(z;s))$ for indicative sample-cluster similarity values $z_1$, $z_2$ and $z_3$, and for varying relative cluster sizes $s$. Note that $s=0$ means the cluster has not been assigned any samples (empty cluster), $s=1$ means the cluster has been assigned all samples (collapse), and $s=\frac{1}{K}$ is the optimal, balanced case.}
    \label{fig:impact_b}
\end{figure}

The impact of $\mathcal{B}$ is illustrated in~\cref{fig:impact_b} for different initial values of $z$ and varying relative cluster sizes $s$. As seen there, for a cluster with optimal size $\frac{1}{K}$, $z^B=\mathcal{B}(z;s)=z$, while for $s<\frac{1}{K}$, $z^B>z$ up to a maximum value of $z^B=1$ for $s=0$, and for $s>\frac{1}{K}$, $z^B<z$ down to a minimum value of $z^B=-1$ for $s=1$. The result of this operation, as explained and demonstrated in~\cref{sec:method} and 4 respectively, is that clusters remain approximately balanced throughout training.

We note here that, as seen in~\cref{eq:b} and~\cref{fig:impact_b}, $\mathcal{B}$ is linear for $s>\frac{1}{K}$. This choice was made because, for $K=65,536$, the range of $[0,\frac{1}{K}]$ is very small relative to $[\frac{1}{K},1]$. We therefore opted for a less steep function in the expectation that it would lead to more stable training. We validate this intuitive choice experimentally in~\cref{tab:abl_b}, where we see that a linear function indeed performs better than an exponential one, although we emphasize that both are stable.

\setlength{\tabcolsep}{10pt}

\begin{table}[t]
\begin{center}
\caption{Linear evaluation accuracy with ResNet-50, pretrained with multi-crop for 100 epochs, with different functions $\mathcal{B}$ for $s<\frac{1}{K}$. The plot demonstrates the corresponding curves of $\mathcal{B}$ for a given $z$ and for varying values of $s$.}
\label{tab:abl_b}
\begin{tabular}{l c c}
\toprule
$\mathbf{B}(z ; s)$ if $s<\frac{1}{K}$ & Linear Acc. & Plot \\
\midrule
 & & \multirow{2}{*}{\includegraphics[width=0.38\linewidth]{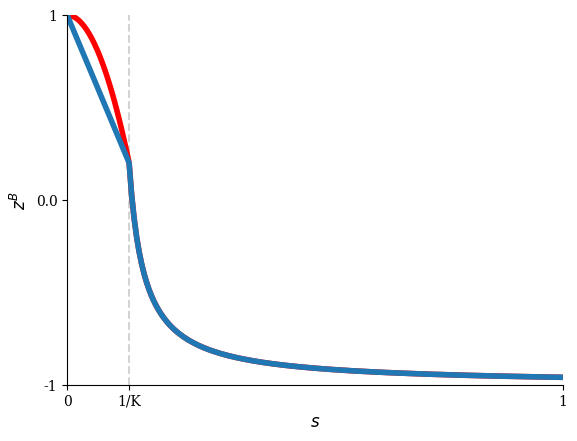}} \\ \\ \\
 $1 - \left[1-z\right](s K)^2$ \hspace{0.1cm} \textcolor{red}{\rule[0.5ex]{0.5cm}{2pt}} & 74.3 \\
  \\
$1 - \left[1-z\right](s K)$ \hspace{0.25cm} \textcolor{blue}{\rule[0.5ex]{0.5cm}{2pt}} & \textbf{74.5} \\ \\ \\ \\
\bottomrule
\end{tabular}
\end{center}
\end{table}

\vspace{5cm}

\section{Embedding Visualization}\label{sec:embedding_viz}

\begin{figure}[t]
    \centering
    \includegraphics[width=0.8\linewidth]{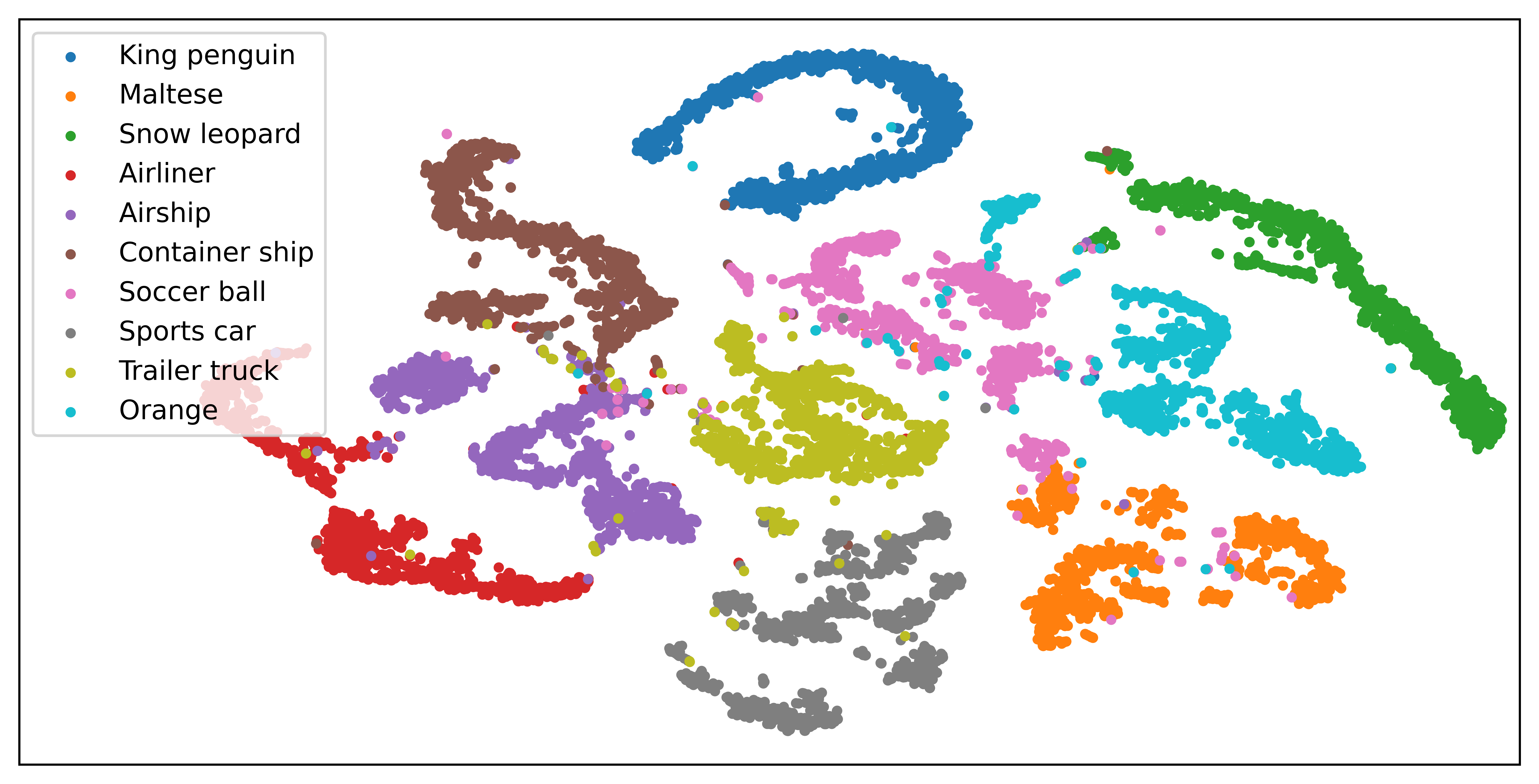}
    \captionsetup{justification=centering} 
    \caption{Embedding visualization for 10 classes in ImageNet's train set.}
    \label{fig:tsne}
\end{figure}

To complement the results presented in~\cref{sec:experiments}, we include a visualization of the features learned by \ours. Specifically we extract features $(h_t\circ f_t)(x)$ from ImageNet's train set for the 10 classes used in the ImageNet-10 subset~\cite{chang2017deep}, and use the t-SNE~\cite{van2008visualizing} algorithm to project them in 2 dimensions.
We present the outcome in~\cref{fig:tsne}, where samples are colour-coded according to their ground truth class label.
We observe that classes are relatively compact to a substantial degree, which is notable given that these embeddings are the result of unsupervised pretraining. 

Further examining~\cref{fig:tsne}, we note that visually and semantically distinct classes, such as "snow leopard" and "king penguin", are well separated from the others.
We observe less clear separation between classes with semantic and visual similarities, such as "container ship", "airship", "airliner" and "trailer truck", which is to be expected, as \ours is entirely unsupervised.
Interestingly, there is also some overlap between the classes "orange" and "soccer ball", which we attribute to the fact that, due to the colour augmentations used during pretraining, self-supervised models (including \ours) learn representations that are strongly colour-invariant and rely to a larger extent on shapes to infer semantic information.

\section{Algorithm}\label{sec:algorithm}

In order to facilitate understanding of \ours, as described in~\cref{sec:method}, we provide pseudo-code for \ours's training steps in Pytorch style in~\cref{alg:training_step}. Distinctly, to illustrate the detach operation applied to centroids for local crops, we present the forward function of the centroid layer $q_s$/$q_t$ in~\cref{alg:qforward}.

\begin{algorithm}
\footnotesize{
\caption{A training step of \ours during pretraining with multi-crop.}
\label{alg:training_step}
{\fontsize{8}{12}\selectfont
\texttt{
\\
\textcolor{olive}{
\# x: A batch of N images \\
\# fs, hs, gs, qs: The student model's components \\
\# gt, ht, qt: The teacher model's components \\ 
\# s: The relative cluster size measuring vector \\ 
\# m, m\_s: The momentum parameters for the model and the balancing operator \\
\# t\_t, t\_s: The temperature parameters for the teacher and student \\ 
\# G, L: The number of global and local views \\
}
\\ \\
x\_G, x\_L = augment\_G(x), augment\_L(x) \textcolor{olive}{\# Global/Local views of samples x} \\
\\
\textcolor{olive}{\# Sample-cluster similarities zh for the projector hs} \\
z\_hG = qs(hs(fs(x\_G)), detach=False) \\ 
z\_hL = qs(hs(fs(x\_L)), detach=True) \\
\\ 
\textcolor{olive}{\# Sample-cluster similarities zg for the predictor gs} \\
z\_gG = qs(gs(hs(fs(x\_G))), detach=True) \\
z\_gL = qs(gs(hs(fs(x\_L))), detach=True) \\ 
\\
\textcolor{olive}{\# Balanced sample-cluster similarities zb for the teacher} \\
with torch.no\_grad(): \\
\-\ \ \ \ \ z\_t = qt(ht(ft(x\_G))) \\
\-\ \ \ \ \ z\_b = balancing(z\_t) \\ 
\\
\textcolor{olive}{\# Cluster probability assignments for the teacher and student} \\
p\_t = softmax(z\_b/t\_t) \# G$\times$N$\times$K \\
p\_h = softmax(cat(z\_hL, z\_hG)/t\_s) \textcolor{olive}{\# G$\times$N$\times$K} \\
p\_g = softmax(cat(z\_gL, z\_gG)/t\_s) \textcolor{olive}{\# L$\times$N$\times$K} \\
\\
\textcolor{olive}{\# Calculation of the loss} \\
loss\_g, loss\_h = 0, 0 \\
for v in range(G): \\
\-\ \ \ \ \ for vg in range(G): \\
\-\ \ \ \ \ \ \ \ \ if v!=v\_g \\
\-\ \ \ \ \ \ \ \ \ \ \ \ \ loss\_h += mean((p\_t[v]*log(p\_h[vg])).sum(dim=1)) \\
\-\ \ \ \ \ \ \ \ \ loss\_g += mean((p\_t[v]*log(p\_g[vg])).sum(dim=1)) \\
\-\ \ \ \ \ for vl in range(L): \\
\-\ \ \ \ \ \ \ \ \ loss\_h += mean((p\_t[v]*log(p\_h[vl])).sum(dim=1)) \\
\-\ \ \ \ \ \ \ \ \ loss\_g += mean((p\_t[v]*log(p\_g[vl])).sum(dim=1)) \\
loss = 0.5 * loss\_h / (G+L-1) + 0.5 * loss\_g / (G+L) \\ 
\\
\textcolor{olive}{\# Student and teacher updates} \\
loss.backward() \\
update(fs, hs, gs, qs) \\
for ms, mt in zip([fs, hs, qs], [ft, ht, qt]): \\
\-\ \ \ \ \ mt = mt * m + ms * (1-m) \\
\\ \\ \\ 
def balancing(z\_t): \\
\-\ \ \ \ \ batch\_labels = z\_t.argmax(dim=-1) \\
\-\ \ \ \ \ s\_b = normalize(one\_hot(batch\_labels), p=1) \\
\-\ \ \ \ \ s = s * m\_s + s\_b * (1-m\_s) \\
\-\ \ \ \ \ z\_b = 1+(1-z\_t)*(1-relu(1-sK)) \textcolor{olive}{\# Efficient implementation of $\mathcal{B}$ for s<1/K} \\
\-\ \ \ \ \ z\_b = (1+z\_b)*(1-relu(1-1/sK))-1 \textcolor{olive}{\# Efficient implementation of $\mathcal{B}$ for s>1/K} \\
\-\ \ \ \ \ return z\_b \\
}
}
}
\end{algorithm}

\begin{algorithm}
\footnotesize{
\caption{The forward function for centroid layers $q$.}
\label{alg:qforward}
{\fontsize{8}{12}\selectfont
\texttt{
\\
\textcolor{olive}{\# K, D, B: The number of clusters, input vector size and batch size} \\
\textcolor{olive}{\# C: The cluster centroids' $K\times D$ matrix}  \\
\textcolor{olive}{\# x: The centroid layer's $B\times D$ input} \\
\\ %
def forward(x, detach=False): \\
\-\ \ \ \ \  xn =  F.normalize(x, p=2, dim=-1) \\
\-\ \ \ \ \  if detach: \\
\-\ \ \ \ \ \ \ \ \  Cn = normalize(C.detach(), p=2, dim=-1) \\
\-\ \ \ \ \  else: \\
\-\ \ \ \ \ \ \ \ \  Cn = normalize(C, p=2, dim=-1) \\
\-\ \ \ \ \  return xn@Cn.T
}
}
}
\end{algorithm}

\end{document}